\documentclass[]{spie}  
\usepackage{graphicx}
\usepackage[T1]{fontenc}
\usepackage{lmodern}
\usepackage{textcomp}
\usepackage{latexsym}
\usepackage{caption}
\usepackage{subfigure}
\usepackage{array}
\usepackage{enumitem}
\usepackage{multirow}

\title{A Retinex-based Image Enhancement Scheme with Noise Aware Shadow-up Function} 

\author{Chien Cheng CHIEN ,Yuma KINOSHITA, Sayaka SHIOTA and Hitoshi KIYA
\skiplinehalf
Tokyo Metropolitan University, 6--6 Asahigaoka, Hino-shi, Tokyo, Japan \\
}
\begin{document} 
  \maketitle 

\begin{abstract}
This paper proposes a novel image contrast enhancement method based on both a noise aware shadow-up function and Retinex (retina and cortex) decomposition. Under low light conditions, images taken by digital cameras have low contrast in dark or bright regions. This is due to a limited dynamic range that imaging sensors have. For this reason, various contrast enhancement methods have been proposed.
Our proposed method can enhance the contrast of images without not only over-enhancement but also noise amplification. In the proposed method, an image is decomposed into illumination layer and reflectance layer based on the retinex theory, and lightness information of the illumination layer is adjusted. A shadow-up function is used for preventing over-enhancement. The proposed mapping function, designed by using a noise aware histogram, allows not only to enhance contrast of dark region, but also to avoid amplifying noise, even under strong noise environments. 
\end{abstract}



\section{INTRODUCTION}

The low dynamic range (LDR) of modern digital cameras is a major factor preventing them from capturing images as well with human vision. This is due to a limited dynamic range which imaging sensors have. For this reason, images taken by digital cameras have low contrast in dark or bright regions. To overcome the problem, various contrast enhancement methods have so far been proposed\cite{KZU,YK,YW,AGCWD,XWU}.

In our previous work\cite{CCH}, we presented a framework of contrast enhancement which can avoid over-enhancement and noise amplification. However, the image enhancement is weak when images include strong noise. Because of this, the result just be enhanced slightly under strong noise environments. The retinex theory allows us to obtain real-world scenes from original images\cite{EDW} such that color is unaffected by reflection, so we use a control factor based on luminance adaptation to reduce the effect of noise on contrast enhancement. We extend our previous work\cite{CCH} by applying the retinex theory to achieve a more naturally and clearly enhanced image, even under strong noise environments.

We evaluate the effectiveness of the proposed method in terms of the quality of enhanced images by a number of simulations. In the simulations, the proposed method is compared with conventional contrast enhancement methods, including state-of-art ones.
Experimental results showed that the proposed method can produce high quality images without over-enhancement, and the proposed method outperforms conventional ones in terms of the noise robustness.

\section{RELATED WORKS} 
\subsection{Retinex theory} 
The basic assumption of Retinex theory can be simply described as $S=R \cdot L$, where original image S is the product of illumination L and reflectance R.
The process of using only a single surround information in the conversion process of each pixel is called Single-Scale Retinex (SSR)\cite{SSR}.
In this process, the Halo artifacts occur unnaturally near the boundary of regions with large gradient.
To solve this problem, Multi-Scale Retinex (MSR)\cite{MSR} was proposed, in which multiple SSR with different areas in the peripheral region are obtained and synthesized by applying appropriate load.
However, by using logarithmic transformation, there was a problem that the result was not stabilized under the influence of noise in dark areas. 
Simultaneous reflection \& illumination estimation (SRIE)\cite{SRIE} and weighted variation model (WVM)\cite{WVM} are Retinex-based methods, in which the illumination is estimated and used for enhanced image combining with Gamma correction, have been proposed.
These methods have a good performances in images without noise, but some strange areas are happened in strong noise environments.
Many outstanding methods\cite{DJJ,LIME,XRE} have been proposed to improve image the quality of images, and preserve more details.

\begin{figure}[t]
\centering
\begin{tabular}{cc}
\begin{minipage}[]{.52\textwidth}
\centering
\includegraphics[width=95mm]{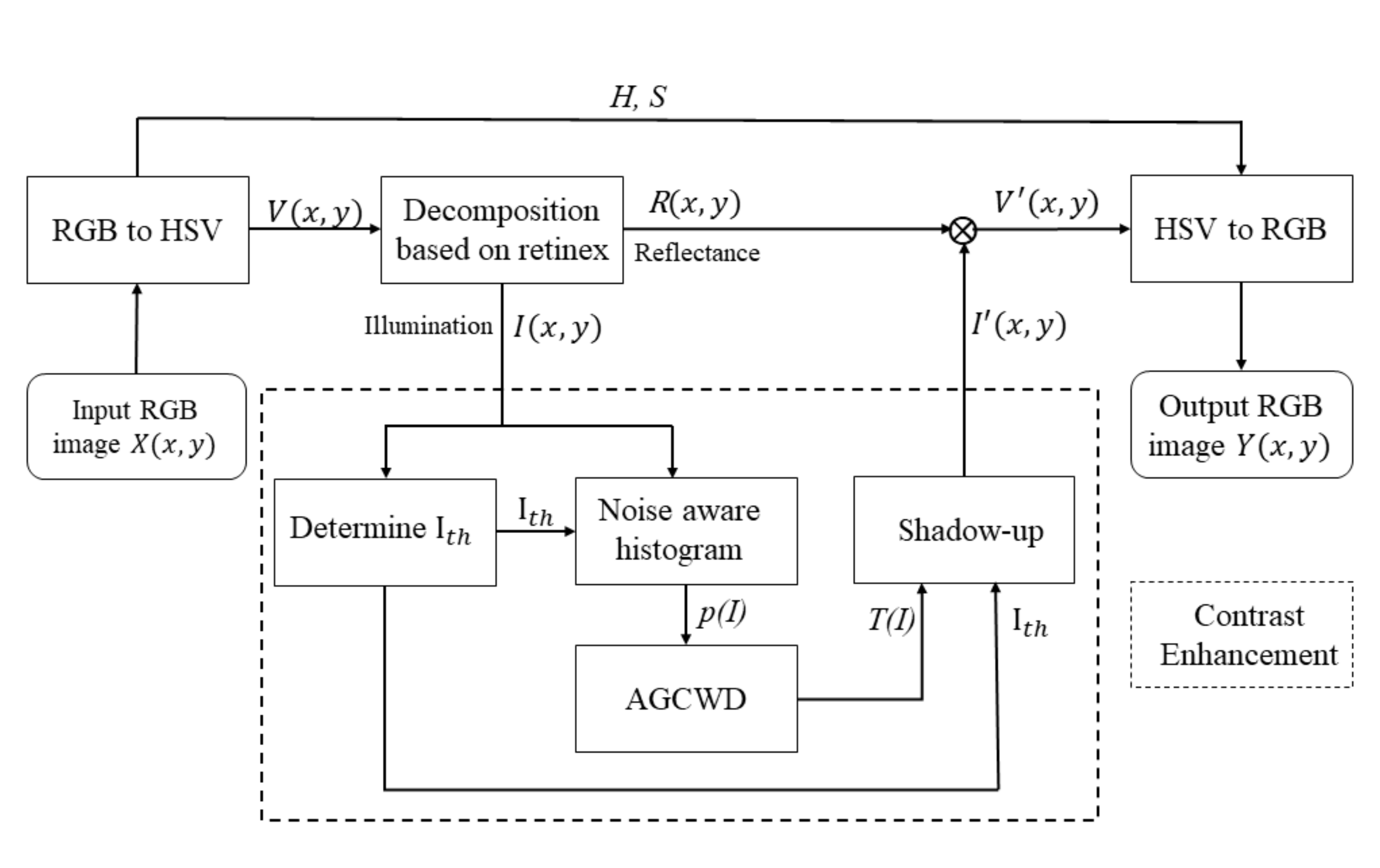}
\caption{Flowchart of the proposed method.}\label{fig:1}
\end{minipage}&
\begin{minipage}[-1cm]{.42\textwidth}
\centering

\subfigure[] {\includegraphics[width=38mm]{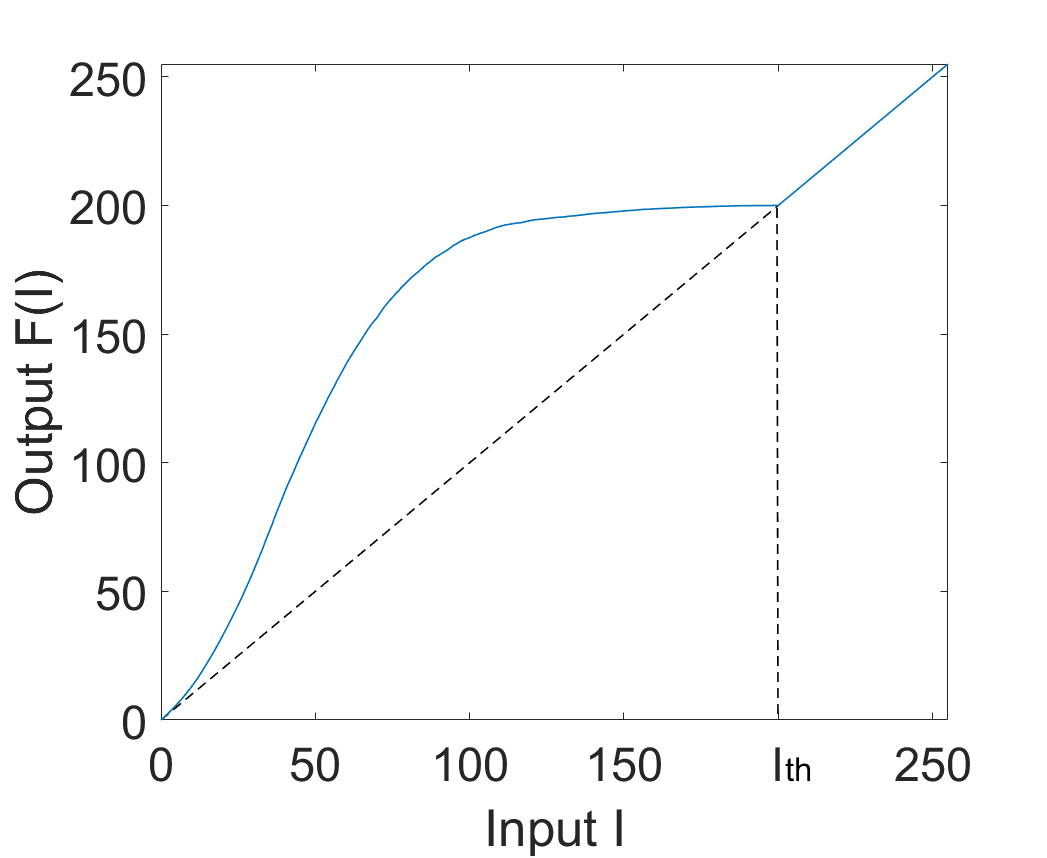}\label{fig:fig02left}
}
\subfigure[] {\includegraphics[width=38mm]{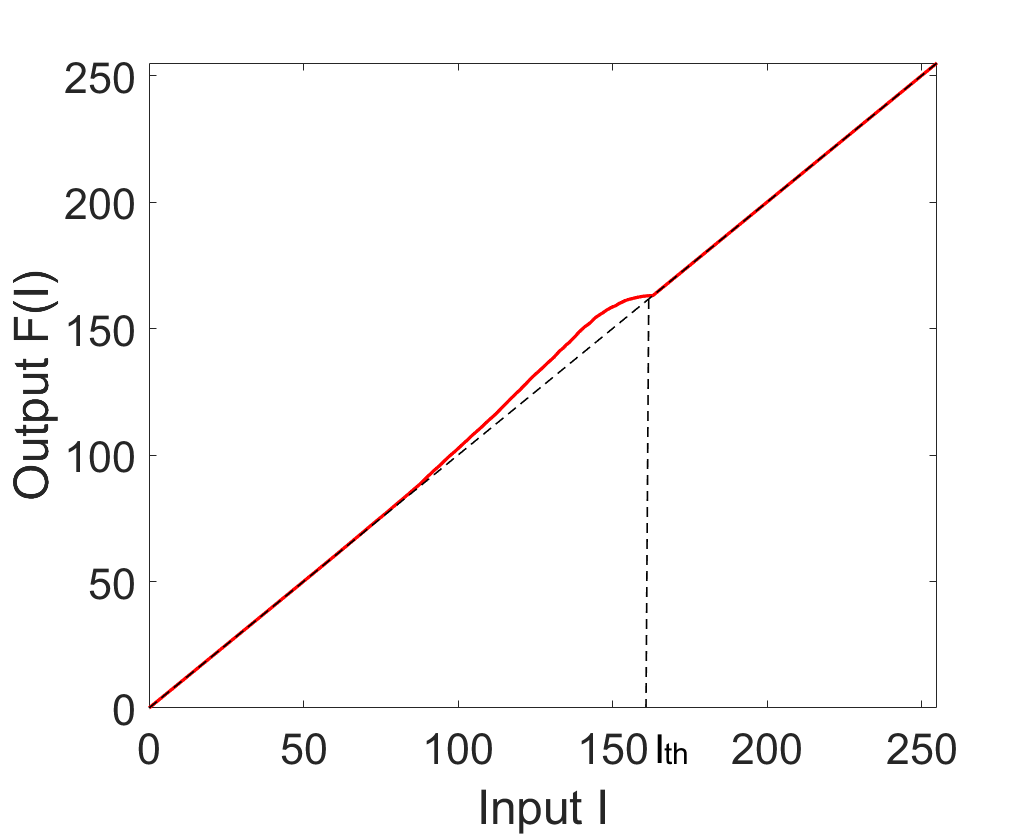}\label{fig:fig02right}
}
\caption{Examples of mapping curves. (a) Mapping curve with shadow-up function. (b) Mapping curve with shadow-up function in noise environments.}\label{fig:2}

\end{minipage}
\end{tabular}
\end{figure}

\subsection{Image enhancement} 
The histogram equalization (HE) \cite{SPI} is one of the most popular algorithms for contrast enhancement\cite{HKS} and various  extended versions of the HE have been proposed \cite{KZU,YK,YW,AGCWD,XWU}. 
Efficient contrast enhancement using adaptive gamma correction with weighting distribution (AGCWD)\cite{AGCWD} aims to prevent over-enhancement and under-enhancement caused by the HE by using an adaptive gamma correction and a modified probability distribution. 
However, there are still issues that the over-enhancement and the loss of contrast in bright areas are caused by these histogram-based methods.
Some noise hidden in the darkness is also amplified. Because of such a situation, some histogram-based contrast enhancement methods preventing the noise amplification have been proposed.
In the methods, shrinkage functions are used for preventing the noise amplification.
Low light image enhancement based on two-step noise suppression (LLIE)\cite{SUH} uses both noise level function (NLF) and just noticeable difference (JND) for contrast enhancement with noise suppression.
Although this method can reduce some noise, it does not preserve the details in bright areas as with histogram-based methods.
Another way for enhancing images is to use multi-exposure image fusion, by using photos with different exposures \cite{MEI, AEC}.

In this paper,we utilize the illumination layer in Retinex theory to reduce the effects of noise and use histogram-based AGCWD to adjust  the illumination layer. We consider not only the noise but also the highlights.
Our purpose enhances contrast with noise suppression, and moreover preserve details in bright regions based on Retinex theory.

\section{PROPOSED SCHEME} 
The outline of the proposed method is shown in Fig \ref{fig:1}.
\subsection{Decomposition based on Retinex} 
As shown in Fig \ref{fig:1}, input RBG image $X(x,y)$ is transformed to an HSV image, where $H(x,y), S(x,y)$ and $V(x,y)$ are Hue, Saturation and Value respectively.
An excellent weighted variational model (WVM) has been proposed\cite{WVM} for simultaneous reflectance and illumination estimation.
We use this model for decomposing V(x,y) into illumination layer $I(x,y)$ and reflectance layer $R(x,y)$, where $I(x,y)$ includes almost no noise, but $R(x,y)$ includes, due to the work of the model.


\subsection{Contrast Enhancement} 
$I(x,y)$ is enhanced by using three key technologies: Shadow-up function, AGCWD, and Noise aware histogram.
The use of a shadow-up function aims to avoid the loss of details in bright areas due to over enhancement.  
Fig \ref{fig:2} illustrates examples of shadow-up functions, which consist of a nonlinear part and a linear part, given by 
\begin{equation}
F(I) = \left\{\begin{array}{ll}
                 T(I)\mbox{, if $I<I_{th}$} \\  
                 I\mbox{, otherwise} \\  
                 \end{array} \right.,
\end{equation}
where $I\in[0, 255]$ is the intensity of illumination layer, $T(I)$ is a monotonically increasing function, and $I_{th}$ is the upper limit for the nonlinear part, which is used for avoiding over enhancement in bright areas.
Contrast is enhanced only when $I$ is less than the threshold value $I_{th}$, according to eq. (1). 
AGCWD is a method to design $T(I)$.
Adaptive gamma correction with AGCWD provides high quality images and it has a low computational cost\cite{AGCWD}.
However, AGCWD usually causes the noise amplification because it does not consider noise characteristics \cite{SUH}. 
To overcome this problem, noise aware histograms and the Retinex theory are applied to AGCWD in this paper.
We compute the noise aware histogram of pixels as follows:
\begin{equation}
p(I)=\frac{\arrowvert B_I \arrowvert}{\arrowvert S \arrowvert},
\end{equation}
where
\begin{eqnarray}
&S=\{(x,y):c(x,y)>n(l(x,y)); l(x,y)<I_{th}\},\\
&B_I=\{(x,y)\in{S}:l(x,y)=I\}.
\end{eqnarray}
$c(x,y)$ is a local contrast estimated by using a Gaussian filter\cite{GEI}, and n(I(x,y)) is the model of noise level\cite{XLI}.
$S$ is the set of pixels having higher contrast than the noise level, and $B_I$ is the subset of $S$ which contains the pixels whose intensity is $I$.

To determine a proper threshold value $I_{th}$ for illumination layer, we take into account the luminance distribution of the illumination layer.
Let $H=\{(x,y):I_{th}<I(x,y)<I_{max}\}$, where $I_{th}$ is the $th$ percentile of luminance $I(x,y)$ of the input image, and $I_{max}$ is the maximum of $I(x,y)$.
The threshold value $I_{th}$ is calculated as follows:
\begin{equation}
I_{th}=255-\frac{1}{\arrowvert H \arrowvert}\sum_{(x,y)\in{H}}I(x,y),
\end{equation}

The threshold value $I_{th}$ become smaller for a brighter image, while $I_{th}$ become larger for a darker image.
\begin{figure*}[!t]
\begin{minipage}[]{.99\textwidth}
\centering
\subfigure[] {
\label{subfigure:orig}
\includegraphics[width=41mm]{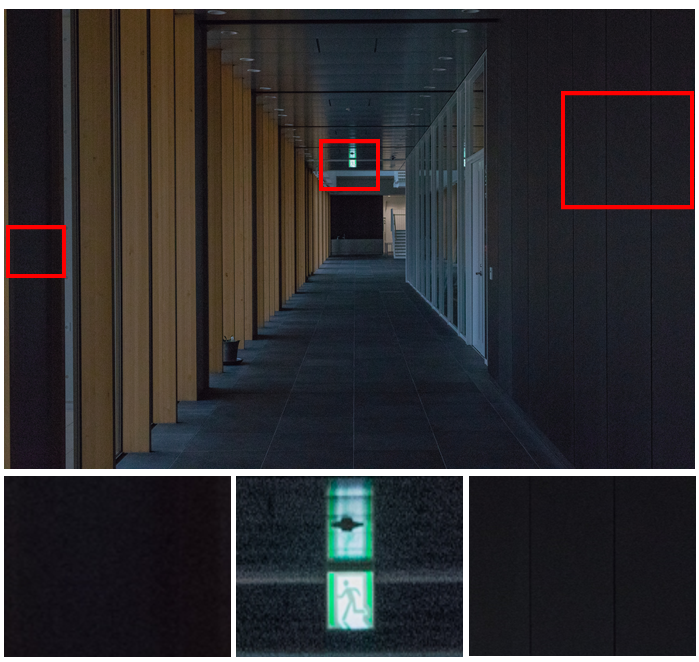}
}
\hspace{-4mm}
\subfigure[] {
\label{subfigure:agc}
\includegraphics[width=41mm]{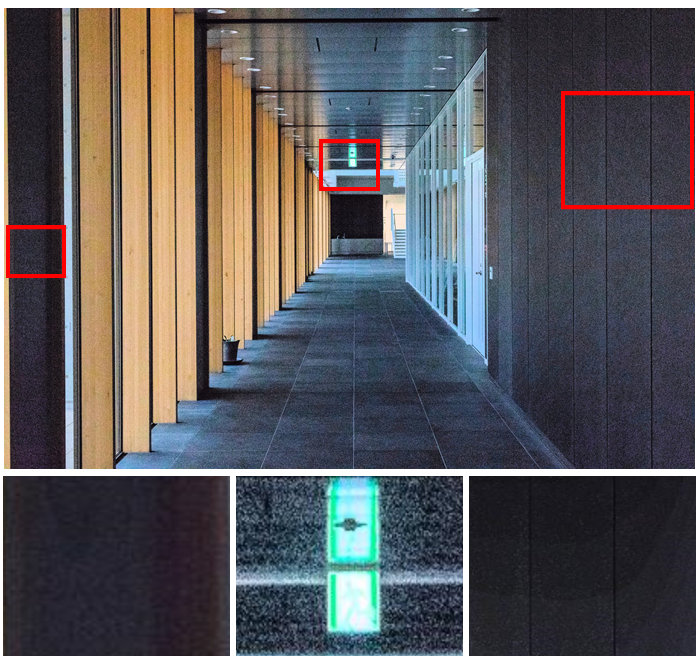}
}
\hspace{-4mm}
\subfigure[] {
\label{subfigure:wvm}
\includegraphics[width=41mm]{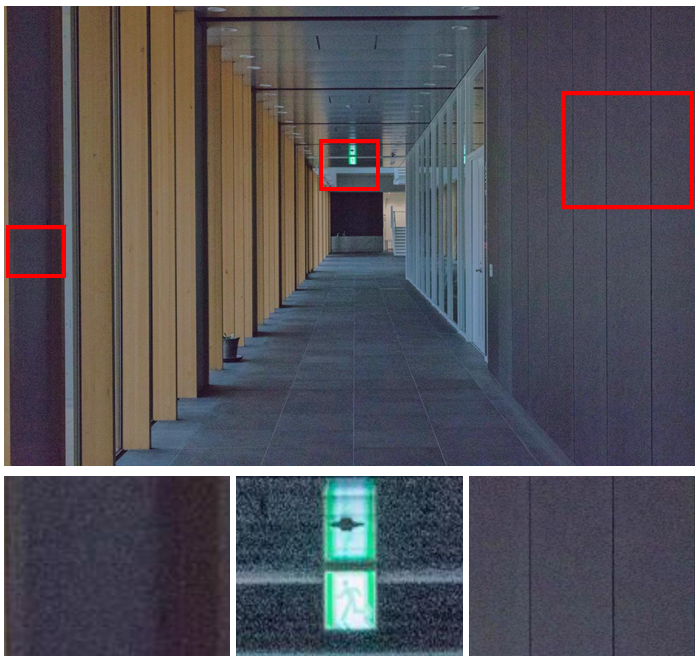}
}
\hspace{-4mm}
\subfigure[] {
\label{subfigure:pro}
\includegraphics[width=41mm]{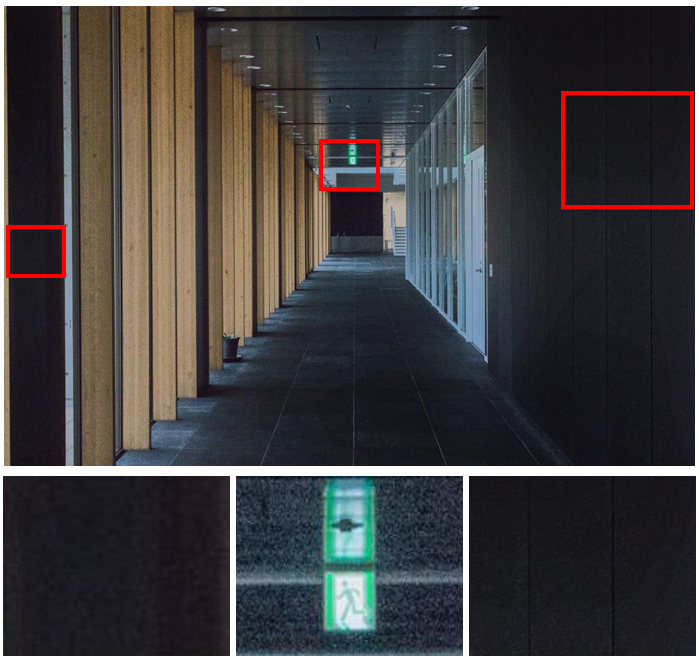}
}
\caption{Experimental Results under different contrast enhancement methods. (a) original image; (b) AGCWD\cite{AGCWD}; (c) WVM\cite{WVM}; (d) Proposed method. The second columns are enlarged views of the previous one.}
\label{fig:3}
\end{minipage}
\begin{minipage}[]{.99\textwidth}
\centering
\subfigure[] {
\label{subfigure:orig_1}
\includegraphics[width=41mm]{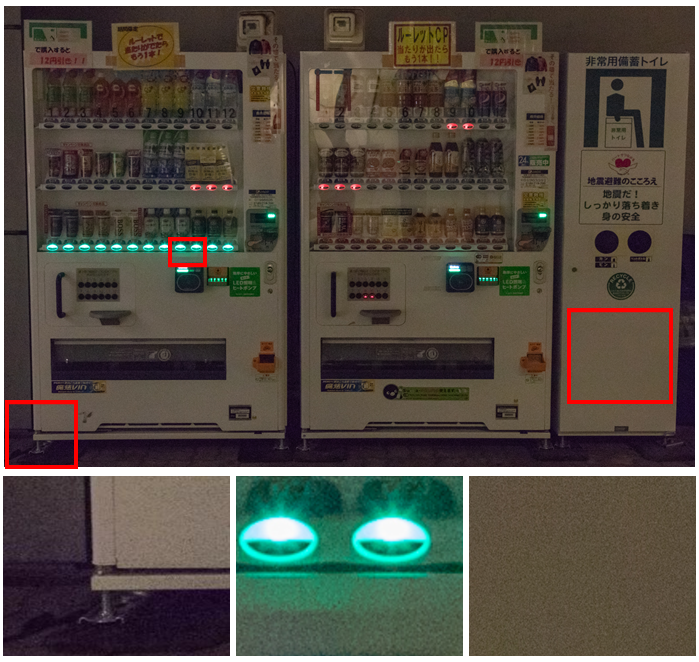}
}
\hspace{-4mm}
\subfigure[] {
\label{subfigure:agc_1}
\includegraphics[width=41mm]{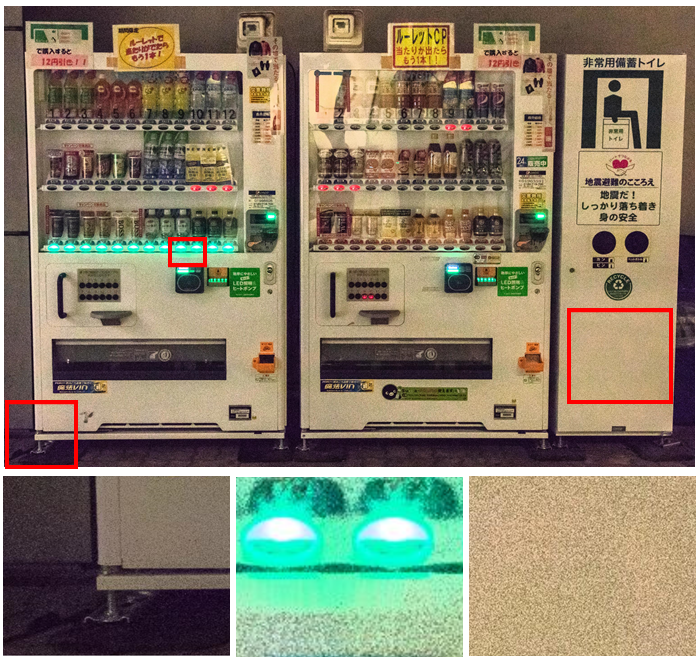}
}
\hspace{-4mm}
\subfigure[] {
\label{subfigure:wvm_1}
\includegraphics[width=41mm]{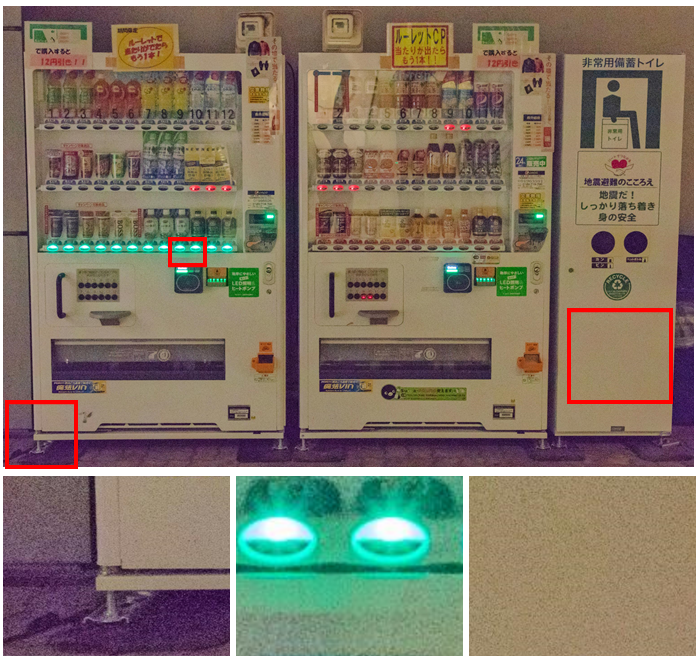}
}
\hspace{-4mm}
\subfigure[] {
\label{subfigure:pro_1}
\includegraphics[width=41mm]{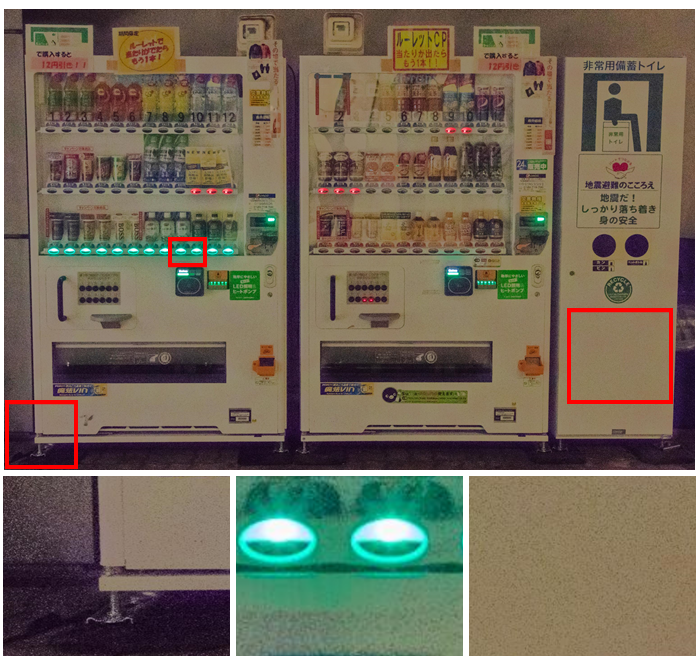}
}
\caption{Experimental Results under different contrast enhancement methods. (a) original image; (b) AGCWD\cite{AGCWD}; (c) WVM\cite{WVM}; (d) Proposed method. The second columns are enlarged views of the previous one.}
\label{fig:4}
\end{minipage}
\end{figure*}


\subsection{Output RGB image $Y(x,y)$} 
By enhancing the illumination layer, an adjusted illumination $I^{\prime}(x,y)$ is obtained. Then an enhanced  $V^{\prime} (x,y)$ is computed by:
\begin{equation}
V^{\prime} (x,y)=I^{\prime}(x,y)\cdot R(x,y)
\end{equation}

Finally, output RGB image $Y(x,y)$ is obtained by transforming $V^{\prime} (x,y), H(x,y)$, and $S(x,y)$, according to the model [14].

\subsection{Proposed procedure} 
The proposed procedure for enhancing an image is summarized as follows (see Fig. 1).
\begin{enumerate}[nosep]
\item Calculate $V(x, y)$ from an input image $X(x, y)$.
\item Decompose $V(x, y)$ into $R(x, y)$ and $I(x, y)$ according to Ref.~\citenum{WVM}.
\item Determine a threshold value $I_{th}$ by eq. (5).
\item Calculate a noise aware histogram $p(I)$ by eq. (2).
\item Design $F(I)$ by using AGCWD\cite{AGCWD} and $I_{th}$.
\item Calculate the enhanced luminance $I^{\prime}(x,y)$ according to eq. (1). 
\item Compute $V^{\prime} (x,y)$ by combining $I^{\prime} (x,y)$ and $R(x,y)$. 
\item Obtain enhanced image $Y(x, y)$ by transforming $H, S$, and $V^{\prime} (x, y)$ into the RGB color space.
\end{enumerate}

\section{SIMULATION} 
\subsection{Simulation condition} 
Images used in this experiment were taken by a digital camera Canon 5D mark IV, where ISO 12800 and safe shutter speed were set.
Because the ISO value was very high, the images contained a lot of noise. The images are shown in Fig \ref{subfigure:orig} and Fig \ref{subfigure:orig_1}.

We carried out a simulation to compare the proposed method with conventional contrast enhancement methods, AGCWD\cite{AGCWD} and WVM\cite{WVM} and LLIE\cite{SUH} for objective evaluation.
The simulations were run on a PC with Intel(R) Core (TM) i7 CPU (3.40GHz) and 8.00GB RAM running a Windows 10 OS and MATLAB 2018a.  We adopted the $75th$ percentile as $I_{th}$.

\subsection{Simulation results} 
\subsubsection{Visual comparison} 
In visual comparison, Fig \ref{subfigure:orig} and Fig \ref{subfigure:orig_1} images were used as input images of each method.
Next, we compared the resulting images, subjectively.
The second columns are enlarged views of the previous column of red box, so that we can see the difference in dark and bright area, also noise.
From Fig \ref{subfigure:agc}), it is confirmed that AGCWD clearly reproduces dark areas and over-emphasizes bright areas. 
Further, we can see noise everywhere obviously and we found significant moire pattern in Fig \ref{subfigure:agc}. 
The same situation occured in Fig \ref{subfigure:agc_1}.
Besides, WVM not only enhanced noise but also white balance was changed in dark area. Also, we easily observed the unusual purple area in Fig \ref{subfigure:wvm} and Fig \ref{subfigure:wvm_1}. 
By comparison the proposed method almost same as the original image in bright area and did not have moire pattern in dark area in Fig \ref{subfigure:pro} and Fig \ref{subfigure:wvm_1}.
To sum up, the proposed method has a better balance between bright areas, dark areas and noise.
\begin{table}[t]
\caption{OBJECTIVE EVALUATION FOR NOISY IMAGES}
\label{table:1}
\begin{center}
\begin{tabular}{c|c|cccc|c}
\hline
\hline

\multicolumn{2}{c|}{Method}&Original&AGCWD\cite{AGCWD}&WVM\cite{WVM}&LLIE\cite{SUH}&Proposed\\ 
\hline
\multirow{2}{*}{NIQE} &Fig 3&6.4461&3.5778&3.1326&4.8199&\boldmath$2.8340$\\ \cline{2-2}

&Fig 4&6.0844&4.1112&2.3942&3.3062&\boldmath$2.1074$\\

\hline
\multirow{2}{*}{ARISM} &Fig 3&3.1113&2.9589&2.8021&2.9979&\boldmath$2.8102$\\ \cline{2-2}

&Fig 4&3.1875&3.1194&2.8897&2.8689&\boldmath$2.8478$\\
\hline
\hline
\end{tabular}
\end{center}
\end{table}

\subsubsection{Objective evaluation} 
A blind image quality assessment called natural image quality evaluator (NIQE)\cite{NIQE} was used to evaluate the enhanced results.
The lower NIQE value represents a higher image quality. As shown in Table \ref{table:1}, our method had a lower value.

Since NIQE is just for naturalness of image assessment, we used another color image assessment called autoregressive- based image sharpness metric (ARISM)\cite{KGU}. 
In Table I, the proposed method had a lower average on NIQE/ARISM than the other state-of-art methods, so that our method is shown to have a good balance and performance in most cases.

\begin{figure}
\begin{center}
\begin{tabular}{c}
\includegraphics[height=5cm]{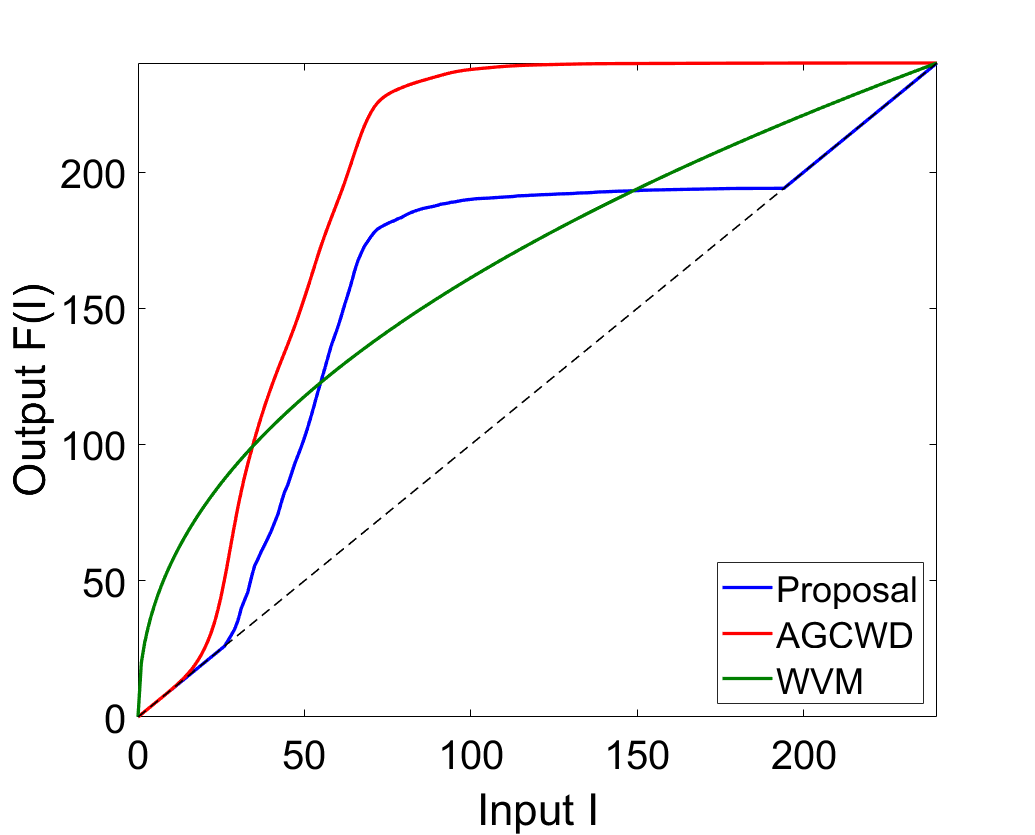}
\end{tabular}
\end{center}
\caption{Mapping curves examples for the image in Fig \ref{fig:3}.}
\label{fig:mapping} 
\end{figure}

Fig \ref{fig:mapping} illustrated the results of mapping curve with each method of Fig \ref{fig:3}. 
In dark area, WVM was too much enhanced.
In bright areas, AGCWD was over-enhanced so it did not preserve details.
In contrast, our proposed method prevented noise amplification in dark areas and preserved more details in bright areas.

\section{CONCLUSION} 
In this paper, we have proposed a novel image contrast enhancement method based on a noise aware shadow-up function.
The proposed method can enhance image contrast without over-enhancement and noise amplification.
To prevent over-enhancement, the proposed method utilizes a Shadow-up function.
In addition, not only the use of noise aware histogram enables us to avoid amplifying noise, but also Retinex theory overcome the image enhancement is weak when images include strong noise.
Experiment results showed that the proposed method successfully enhances contrast while preserving details of highlight regions and suppressing some noise in dark regions.


\end{document}